\documentclass{article}
\usepackage{spconf,amsmath,graphicx}
\usepackage{amssymb,amsmath}

\usepackage{multirow,multicol}
\usepackage{booktabs}
\usepackage{makecell,rotating}
\usepackage{pifont}

\newcommand{\figref}[1]{Figure \ref{#1}}
\newcommand{\tabref}[1]{Table \ref{#1}}

\usepackage[dvipsnames]{xcolor}
\usepackage{color}
\definecolor{myRed}{RGB}{195,40,40}
\definecolor{myGreen}{RGB}{0,102,0}

\newcommand{\upmark}{\textcolor{myRed}{$\uparrow$}}

\usepackage{algorithm}
\usepackage{algorithmicx}
\usepackage{algpseudocode}

\usepackage{pifont}

\title{Mixed Sample Augmentation for Online Distillation}

\name{
Yiqing Shen\textsuperscript{1,2*,$\dag$}, Liwu Xu\textsuperscript{2}, Yuzhe Yang\textsuperscript{2}, Yaqian Li\textsuperscript{2*}, Yandong Guo\textsuperscript{2}
\thanks{
* Contact Email :yshen92@jhu.edu, liyaqian@oppo.com. }
\thanks{$\dag$ Work done during the internship at OPPO Research Institute.}}
\address{
\textsuperscript{1} 
Shanghai Jiao Tong University, Shanghai, China
\\ 
\textsuperscript{2} OPPO Research Institute, Shanghai, China
}
\begin{document}
%
\maketitle
\begin{abstract}
Mixed Sample Regularization (MSR), such as MixUp or CutMix, is a powerful data augmentation strategy to generalize convolutional neural networks. Previous empirical analysis has illustrated an orthogonal performance gain between MSR and conventional offline Knowledge Distillation (KD). To be more specific, student networks can be enhanced with the involvement of MSR in the training stage of sequential distillation. Yet, the interplay between MSR and online knowledge distillation, where an ensemble of peer students learn mutually from each other, remains unexplored. To bridge the gap, we make the first attempt at incorporating CutMix into online distillation, where we empirically observe a significant improvement. Encouraged by this fact, we propose an even stronger MSR specifically for online distillation, named as Cut\textsuperscript{n}Mix. Furthermore, a novel online distillation framework is designed upon Cut\textsuperscript{n}Mix, to enhance the distillation with feature level mutual learning and a self-ensemble teacher. Comprehensive evaluations on CIFAR10 and CIFAR100 with six network architectures show that our approach can consistently outperform state-of-the-art distillation methods. 
\end{abstract}

\section{Introduction}
Despite the promising achievements of convolutional neural networks (CNNs) achieved in various tasks, how to effectively improve the generalization ability remains to be a preliminary research issue. Previous works have made various attempts in regularization schemes such as dropout \cite{dropout}, label smoothing regularization \cite{lsr}. Knowledge Distillation (KD) \cite{kd} also demonstrates a bright promise to generalize networks by providing sample-wise label smoothing \cite{kd_beyond,tf-kd}. Specifically, dark knowledge is transferred to a lightweight model by forcing it to mimic the soft labels generated from a cumbersome static teacher model. Consequently, the overall objective function that incorporates an additional distillation loss is defined as 
\begin{equation}
    \mathcal{L}_{KD} = \tau^2 \cdot  \mathcal{D}_{KL}(\sigma(\frac{\mathbf{z}^s}{\tau}),\sigma(\frac{\mathbf{z}^t}{\tau})),
\end{equation}
where $\tau$ is the temperature, $\sigma$ is the softmax operation, $\mathcal{D}_{KL}$ is the Kullback-Leibler (KL) divergence, $\mathbf{z}^s$ and $\mathbf{z}^t$ represent the logits from student and teacher respectively. Many works follow this line to generalize a single network by self regularizing the dark knowledge \cite{cs-kd,ps-kd}.

Another popular line to boost generalization ability concentrates on the data augmentation strategy, such as random crop and horizontal flip. More powerful Mixed Sample Augmentation (MSA) strategies, including MixUp \cite{mixup} and CutMix \cite{cutmix}, provide a new outlook on the concept of the vicinity. Concretely, two different training images and their associated labels are linearly interpolated in Mixup to boost the training sample diversity \cite{mixup}. As a combination of Mixup \cite{mixup} and CutOut \cite{cutout}, CutMix cuts and swaps two images $\mathbf{x}_A,\mathbf{x}_b\in\mathbb{R}^{C\times W\times H}$ to derive an intermediate sample:
\begin{equation}
    \widetilde{\mathbf{x}} = \mathbf{M} \odot \mathbf{x}_A + (1 - \mathbf{M}) \odot \mathbf{x}_B, \label{eq:cutmix1}
\end{equation}
where $\odot$ denotes the pixel-wise multiplication and mask $\mathbf{M}\in \{0,1\}^{W\times H}$. Meanwhile, the one-hot labels $y_A,y_B$ are mixed in the same fashion i.e.
\begin{equation}
    \widetilde{y} = \lambda y_A + (1-\lambda) y_b \label{eq:cutmix2}
\end{equation}
with a random balancing coefficient $\lambda$ sampled from beta distribution $Beta(1,1)$ \cite{cutmix}.

\begin{figure}[!t]
\centering
\includegraphics[width=1.0\linewidth]{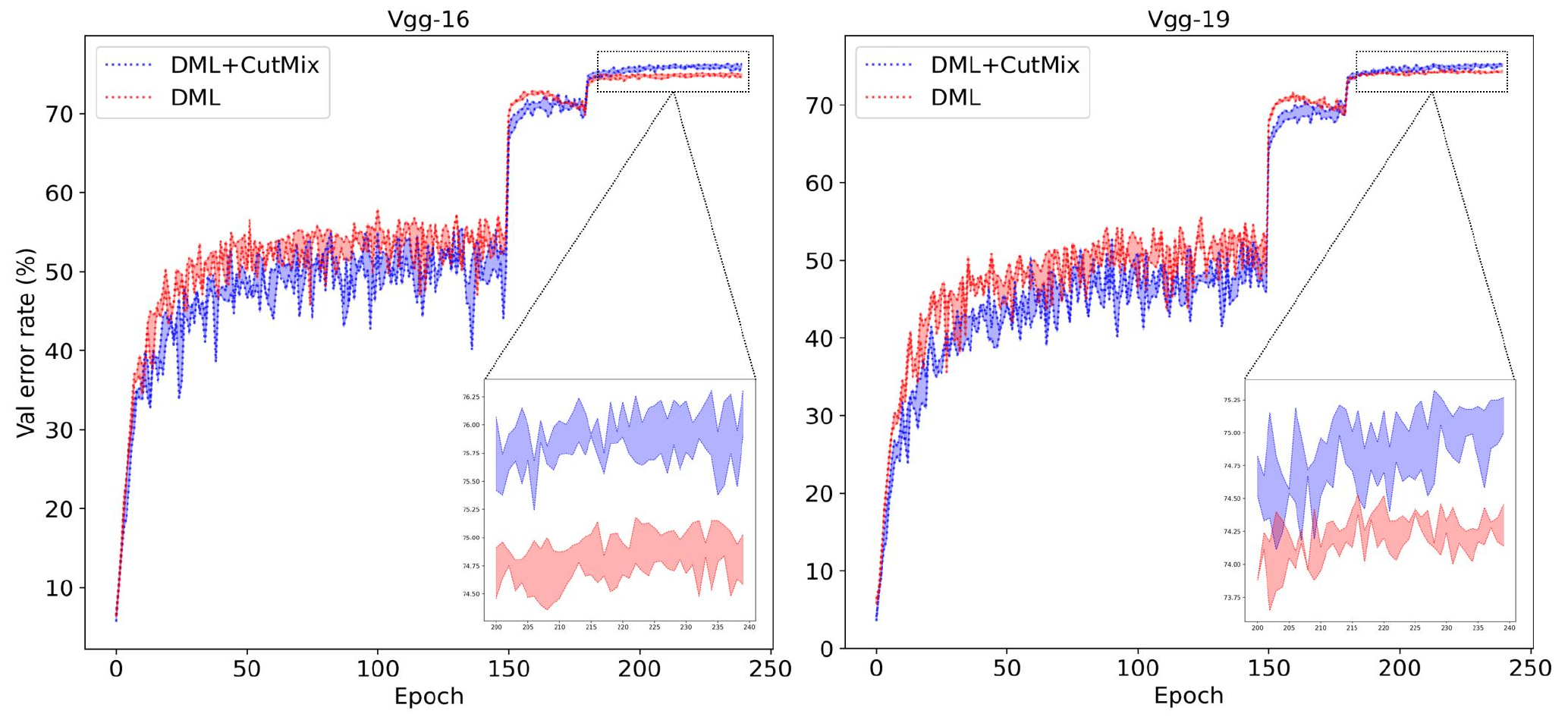}
\caption{Impact of plugging CutMix \cite{cutmix} to DML \cite{dml} on CIFAR-100. Each result is computed by averaging 3 random runs, where the shaded area indicates the standard deviation. A marginal performance improvement in peer students can be observed with CutMix.} 
\label{fig:0}
\end{figure}

\begin{figure*}[!htbp]
\centering
\includegraphics[width=0.8\linewidth]{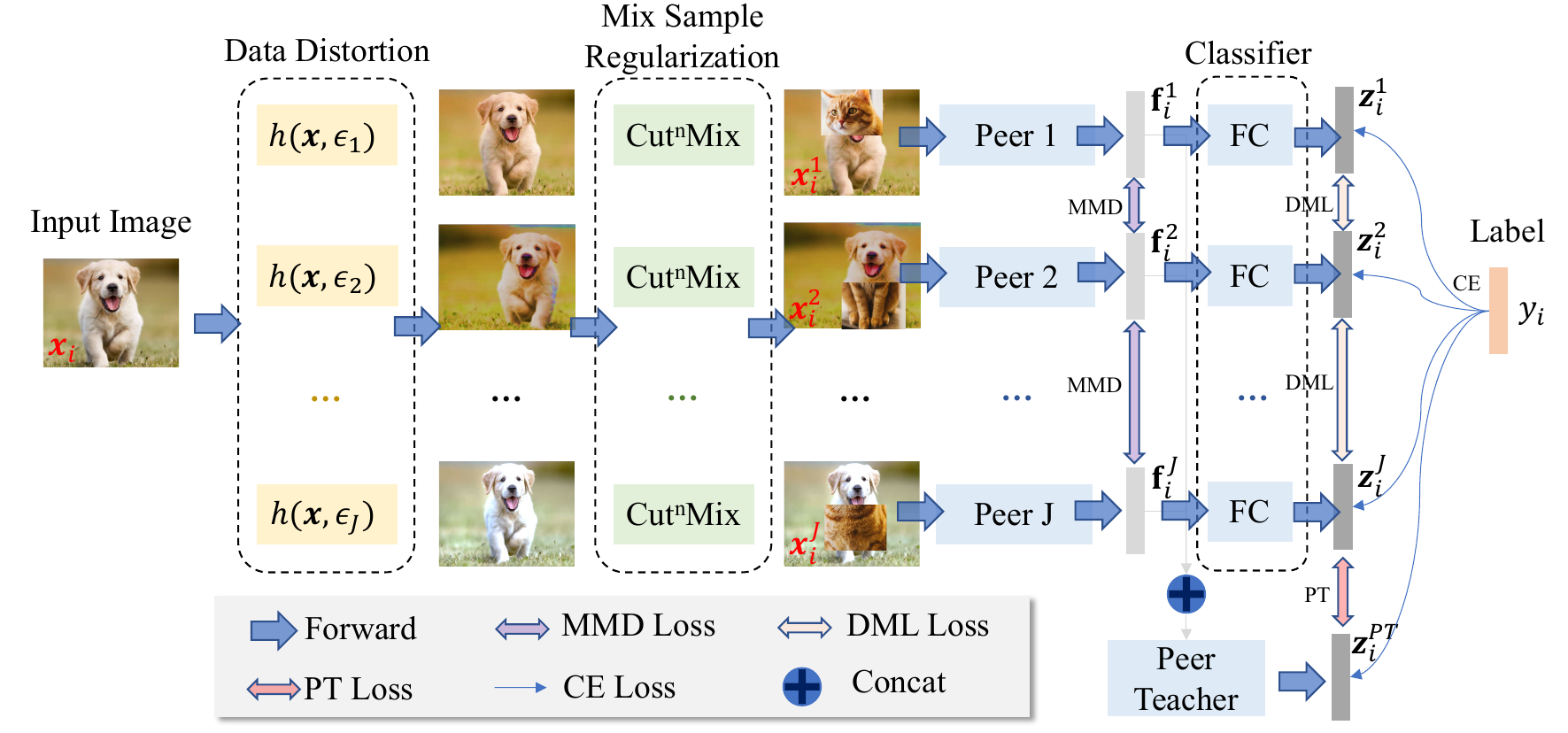}
\caption{The overall framework of the proposed online distillation and Cut\textsuperscript{n}Mix.}
\label{fig:1}
\end{figure*}

It follows that one would naturally expect an ideal property named "orthogonality" \cite{orthogonality} where performance improvement achieved by KD does not conflict with MSA. Previous works have empirically observed an interplay between the conventional two-stage KD and data augmentation including MSAs. Specifically, training the student network with MSA witnesses a significant performance gain \cite{kd_mix}. Whilst, the student is impaired if its associated teacher is trained with any MSA schemes \cite{kd_mix2}. Yet, the compatibility between MSA and one-stage online distillation \cite{dml}, where an ensemble of student networks learns mutual from each other, has not been fully explored. Compared with two-stage distillation, online distillation alleviates the limitation of heavy training time and expensive computational cost, since no complex pre-trained model is required \cite{one,pcl,okddip}. Consequently, exploring the compatibility between online distillation and MSR is an essential research topic. In this work, we primarily focus on CutMix, which has demonstrated superiority among other MSRs when it is combined with the conventional two-stage KDs.

The major contributions are four-fold, summarized as follows. 1) To bridge the gap between MSR and online distillation, we make the first attempt to empirically uncover an interesting phenomenon that MSR can consistently improve deep mutual learning \cite{dml}. 2) Inspired by this fact, we propose an even stronger online distillation-specific MSR strategy, named Cut\textsuperscript{n}Mix. 3) Moreover, we build a novel online distillation framework upon Cut\textsuperscript{n}Mix to achieve more significant performance gain. 4) Finally, we conduct comprehensive experiments on CIFAR-10 and CIFAR-100 to show that our method can outperform state-of-the-art KD methods. 

\section{Methods}
\subsection{Preliminaries}
We focus on the classification task, given $N$ annotated samples $\{(\textbf{x}_i,y_i\}_{i=1}^N$ from $K$ classes. In online distillation \cite{dml}, multiple peer students learn from each other. Specifically, when given $J$ networks, then the objective function of $j$-th network is
\begin{equation}
\begin{aligned}
    \mathcal{L}^j &= \mathcal{L}_{CE}^j + \alpha \cdot \tau^2 \cdot \mathcal{L}_{DML}^j \\ 
    & \text{with} ~ \mathcal{L}_{DML}^j = \frac{1}{J-1}\sum_{k \not= j} \mathcal{D}_{DML}(\sigma(\frac{\mathbf{z}^j}{\tau}),\sigma(\frac{\mathbf{z}^k}{\tau})),\label{eq:dml}
\end{aligned}
\end{equation}
where $\mathcal{L}_{CE}^j$ is the cross-entropy loss, $\textbf{z}^j$ is the logits from $j$-th network, and the balancing coefficient $\alpha$ is set to 1 in \cite{dml}. Subsequently, we plug the CutMix into DML by assigning identically the same batch of images after mixed operation to each network. To be more specific, a batch of samples is augmented progressively by Eq.\eqref{eq:cutmix1}-\eqref{eq:cutmix2} to derive uniform distorted samples from the peer students. The combination of CutMix and DML results in a marginal performance improvement, as illustrated in \figref{fig:0}. Inspired by this fact, we propose a more powerful online distillation with mixed sample augmentation, as depicted in \figref{fig:1}. 

\subsection{Cut\textsuperscript{n}Mix}
We adopt data distortion with different random seeds for raw images separately to each network, which targets to enhance the invariance against perturbations in the data domain \cite{kdcl}. We denote the batch of augmented images associated with $j$-th network as $\mathcal{B}^j = \{\textbf{x}_i^j,y_i\}_{i=1}^n$, where $n$ writes for the batch size, and the superscript $j$ is omitted for $y_i$ as all versions of distorted batches share the identical ground truth labels. Subsequently, in our Cut\textsuperscript{n}Mix, the mixed training sample and the label are generated as follows:
\begin{align}
    \widetilde{\textbf{x}}_i^j  & = \mathbf{M}^j \odot \mathbf{x}_k^j + (1 - \mathbf{M}^j) \odot \mathbf{x}_i^j, \label{eq:mix1}\\
    \widetilde{y}_i^j &= \lambda^j \cdot y_k + (1-\lambda^j) \cdot  y_i \label{eq:mix2}.
\end{align}

Each binary masks $\textbf{M}^j$ indicates
where to drop out and fill in from two images following the previous sampling strategy in \cite{cutmix}. Notably, we impose an additional constraint on the combination ratio as follows
\begin{equation}
    \lambda^j = \lambda ~(\text{for}~ j = 1,2,\cdots,J),\label{eq:constraint}
\end{equation}
where $\lambda$ is sampled from beta distribution $Beta(1,1)$. Eq.\eqref{eq:constraint} guarantees a valid online distillation process, since each network is trained with identical series of ground truths.   

\subsection{Online Distillation with Cut\textsuperscript{n}Mix}
To facilitate the online distillation on more variant data samples augmented by Cut\textsuperscript{n}Mix, we adopt a Maximum Mean Discrepancy (MMD) loss \cite{mmd} to assist the feature level distillation and employ a classifier $g$ as a peer teacher by assembling the feature representations from penultimate layers \cite{pcl}. Subsequently, the overall training optimization objective for $j$-th network is made up of four components: 1) standard cross-entropy loss $\mathcal{L}_{CE}^j$, 2) logit-level mutual distillation loss $\mathcal{L}_{DML}^j$ in Eq.\eqref{eq:dml} to learn from soft labels, 3) MMD loss $\mathcal{L}_{MMD}^j$ for peer feature-level distillation, and 4) peer teacher distillation loss $\mathcal{L}_{PT}^j$ to transfer knowledge from the capable assembled teacher. Thus, the overall loss function is 
\begin{equation}
    \mathcal{L}^j = \mathcal{L}_{CE}^j + \alpha \cdot \mathcal{L}_{DML}^j + \beta \cdot \mathcal{L}_{MMD}^j + \gamma \cdot \mathcal{L}_{PT}^j,\label{eq:loss_all}
\end{equation}
with three balancing coefficient $\alpha,\beta,\gamma$, for $j=1,2,\cdots,J$.

We write the feature representations generated from $\textbf{x}^j_i$ by the $j$-th network as $\textbf{f}_i^j$. Then the empirical MMD loss, targeting feature level mutual distillation, is formulated as:
\begin{equation}
    \mathcal{L}_{MMD}^j =\frac{1}{J-1} \cdot \sum_{k\not=j} \|\frac{1}{n}\sum_{i=1}^n \textbf{f}_i^j -\frac{1}{n}\sum_{i=1}^n \textbf{f}_i^k  \|_2^2, \label{eq:mmd}
\end{equation}
where $\|\cdot\|_2$ is the $L$-2 norm. Peer teacher is a classifier, using the feature representations from peer networks i.e. $\textbf{z}_i^{PT} = g(\textbf{f}_i^1,\cdots,\textbf{f}_i^J)$, which is trained by a standard cross-entropy loss. Then, we transfer the knowledge from the ensemble teacher to
peer student by:
\begin{equation}
    \mathcal{L}_{PT}^j = \tau^2 \cdot \frac{1}{n}\sum_{i=1}^n \mathcal{D}_{KL}(\sigma(\frac{\mathbf{z}_i^j}{\tau}),\sigma(\frac{\mathbf{z}_i^{PT}}{\tau})).\label{eq:pt_kd}
\end{equation}
We summarize the overall training process in Algorithm \ref{alg}. 

\begin{algorithm}[htbp!]
\caption{Proposed Online Distillation with Cut\textsuperscript{n}Mix} \label{alg}
\begin{algorithmic}[1]
\Require Training set $\{(\textbf{x}_i,y_i)\}_{i=1}^N$
\For{i = 1, $\cdots$, Max\_Epoch}
\State Random sample a batch of data $\{(\textbf{x}_i,y_i)\}_{i=1}^n$
\State Derive the distorted data $\{(\textbf{x}_i^j,y_i)\}_{i=1}^n$
\State Mix the data and label by Eq. \eqref{eq:mix1}, \eqref{eq:mix2} 
\State Compute features and logits
\State Assemble features by the peer teacher
\State Compute cross entropy loss $\mathcal{L}_{CE}^j$
\State Compute distillation loss $\mathcal{L}_{DML}^j$ in Eq. \eqref{eq:dml}
\State Compute MMD loss $\mathcal{L}_{MMD}^j$ in Eq. \eqref{eq:mmd}
\State Compute peer teacher distillation loss $\mathcal{L}_{PT}^j$ in Eq. \eqref{eq:pt_kd}, and peer teacher supervision loss
\State Update peer students by Eq. \eqref{eq:loss_all}
\State Update peer teacher
\EndFor
\end{algorithmic}
\end{algorithm}

\section{Experiments}

\begin{table*}[!ht]
\caption{
Top-1 classification accuracy(\%) comparison on CIFAR-10 and CIFAR-100 with the state-of-the-arts. The best and second best performance are highlighted in \textcolor{myRed}{\textbf{Red}} and \textcolor{myGreen}{Green} respectively.
}\label{table:result}
\centering
\resizebox{0.9\linewidth}{!}{
\begin{tabular}{c|l|cccccc} 
\toprule
Dataset&Methods & Vgg-16 & Vgg-19 & ResNet-32 & ResNet-110 & WideResNet & DenseNet \\
\hline 
\multirow{10}{*}{CIFAR-10}
& Baseline &  93.89\tiny{$\pm$0.06} & 93.93\tiny{$\pm$0.06} & 93.51\tiny{$\pm$0.05} & 94.83\tiny{$\pm$0.32} & 94.47\tiny{$\pm$0.10} & 92.89\tiny{$\pm$0.28} \\
& DML \cite{dml} & 94.13\tiny{$\pm$0.37} & 93.93\tiny{$\pm$0.18} & 93.91\tiny{$\pm$0.11} & 94.46\tiny{$\pm$0.24} & \textcolor{myGreen}{95.34\tiny{$\pm$0.05}} & \textcolor{myGreen}{93.26\tiny{$\pm$0.33}} \\
& CutMix \cite{cutmix}  & 94.71\tiny{$\pm$0.03} &94.59\tiny{$\pm$0.02} &94.04\tiny{$\pm0.20$} &95.18\tiny{$\pm$0.07} & 95.11\tiny{$\pm$0.22} & 93.24\tiny{$\pm$0.12} \\
\cline{2-8}
& CS-KD \cite{cs-kd} & 93.72\tiny{$\pm$0.20} & 93.54\tiny{$\pm$0.04} & 93.21\tiny{$\pm$0.18} & 93.90\tiny{$\pm$0.05} & 95.03\tiny{$\pm$0.23} & 92.13\tiny{$\pm$0.27}\\
& DDGSD \cite{r-aug} & \textcolor{myGreen}{94.35\tiny{$\pm$0.14}} & \textcolor{myGreen}{94.27\tiny{$\pm$0.08}} & \textcolor{myGreen}{94.05\tiny{$\pm$0.04}} & \textcolor{myGreen}{95.24\tiny{$\pm$0.04}} & 95.29\tiny{$\pm$0.13} & 93.18\tiny{$\pm$0.21} \\
& KDCL \cite{kdcl} & 93.67\tiny{$\pm$0.09} & 93.49\tiny{$\pm$0.08} & 93.71\tiny{$\pm$0.02} & 94.91\tiny{$\pm$0.17} & 94.57\tiny{$\pm$0.10} & 92.99\tiny{$\pm$0.17} \\
& ONE \cite{one} & 93.70\tiny{$\pm$0.29} & 93.73\tiny{$\pm$0.13} & 94.03\tiny{$\pm$0.06} & 95.24\tiny{$\pm$0.16} & 94.54\tiny{$\pm$0.09} & 92.92\tiny{$\pm$0.43} \\
& OKDDip \cite{okddip} & 93.57\tiny{$\pm$0.18} & 93.33\tiny{$\pm$0.25} & 93.55\tiny{$\pm$0.03} & 95.21\tiny{$\pm$0.25} & 94.47\tiny{$\pm$0.18} & 92.28\tiny{$\pm$0.06} \\
\cline{2-8}

& Ours & \textcolor{myRed}{\textbf{94.84}\tiny{$\pm$0.25}} & \textcolor{myRed}{\textbf{94.81}\tiny{$\pm$0.06}} & \textcolor{myRed}{\textbf{94.24}\tiny{$\pm$0.04}} & \textcolor{myRed}{\textbf{95.76}\tiny{$\pm$0.14}} & \textcolor{myRed}{\textbf{95.90}\tiny{$\pm$0.01}} & \textcolor{myRed}{\textbf{93.64}\tiny{$\pm$0.12}}     \\
& & \small(0.95 \upmark) & \small(0.88 \upmark) & \small(0.73 \upmark) & \small(0.93 \upmark) & \small(1.48 \upmark) & \small(0.75 \upmark) \\
\midrule


\multirow{10}{*}{CIFAR-100}
& Baseline &  73.72\tiny{$\pm$0.15} & 72.83\tiny{$\pm$0.59} & 71.71\tiny{$\pm$0.24} & 76.28\tiny{$\pm$0.29} & 77.71\tiny{$\pm$0.05} & 71.62\tiny{$\pm$0.47} \\
& DML \cite{dml} & 75.39\tiny{$\pm$0.13} & 74.14\tiny{$\pm$0.03} & 72.66\tiny{$\pm$0.03} & 77.90\tiny{$\pm$0.01} & \textcolor{myGreen}{79.57\tiny{$\pm$0.05}} & \textcolor{myGreen}{72.36\tiny{$\pm$0.05}} \\
& CutMix \cite{cutmix}  & 75.43\tiny{$\pm$0.22} &74.34\tiny{$\pm$0.03} &72.69\tiny{$\pm0.30$} &78.58\tiny{$\pm$0.08} & 79.43\tiny{$\pm$0.19} & 71.74\tiny{$\pm$0.30} \\
\cline{2-8}

& CS-KD \cite{cs-kd} & 74.53\tiny{$\pm$0.30} & 73.59\tiny{$\pm$0.70} &
70.85\tiny{$\pm$0.33} & 76.79\tiny{$\pm$0.24} & 78.39\tiny{$\pm$0.20} & 70.50\tiny{$\pm$0.40} \\
& DDGSD \cite{r-aug} & \textcolor{myGreen}{75.65\tiny{$\pm$0.04}} & \textcolor{myGreen}{75.32\tiny{$\pm$0.01}} & 73.68\tiny{$\pm$0.23} & 77.45\tiny{$\pm$0.17} & 79.17\tiny{$\pm$0.08} & 72.32\tiny{$\pm$0.32} \\

& KDCL \cite{kdcl} & 73.04\tiny{$\pm$0.04} & 72.19\tiny{$\pm$0.07} & 72.26\tiny{$\pm$0.04} & 78.37\tiny{$\pm$0.22} & 79.49\tiny{$\pm$0.01} & 72.16\tiny{$\pm$0.71} \\
& ONE \cite{one} & 73.33\tiny{$\pm$0.03} & 72.09\tiny{$\pm$0.02} & \textcolor{myGreen}{73.71\tiny{$\pm$0.93}} & \textcolor{myGreen}{78.91\tiny{$\pm$0.19}} & 78.62\tiny{$\pm$0.24} & 71.45\tiny{$\pm$0.02} \\
& OKDDip \cite{okddip} & 73.97\tiny{$\pm$0.07} & 71.71\tiny{$\pm$0.14} & 72.34\tiny{$\pm$0.06} & 78.17\tiny{$\pm$0.16} & 79.34\tiny{$\pm$0.20} & 70.62\tiny{$\pm$0.01} \\
\cline{2-8}

& Ours & \textcolor{myRed}{\textbf{76.56}\tiny{$\pm$0.28}} & \textcolor{myRed}{\textbf{75.68}\tiny{$\pm$0.20}} & \textcolor{myRed}{\textbf{73.84}\tiny{$\pm$0.25}} &  \textcolor{myRed}{\textbf{79.32}\tiny{$\pm$0.35}} & \textcolor{myRed}{\textbf{80.09}\tiny{$\pm$0.07}} & \textcolor{myRed}{\textbf{72.64}\tiny{$\pm$0.12}}\\
& & \small(2.84 \upmark) & \small(2.75 \upmark) & \small(2.13 \upmark) & \small(3.04\upmark) & \small(2.38 \upmark) & \small(1.02 \upmark) \\
\bottomrule
\end{tabular}
}
\end{table*}

\subsection{Datasets and Implementations}
We evaluate the proposed method on two commonly used image classification datasets, namely CIFAR-10 and CIFAR-100 \cite{cifar}. Each set contains a total number of 60,000 RGB natural images scaled at $32\times32$ pixels from 10/100 classes, where 50,000 images are set aside for training and the rest 10,000 for testing. We employ six popular CNN backbones for performance comparison, including Vgg-16, Vgg-19 \cite{vgg}, ResNet-18, ResNet-110 \cite{resnet}, WideResNet-20-8 \cite{wrn}, and DenseNet-40-12 \cite{densenet}.  

All experiments are carried out on NVIDIA Tesla V100 GPU with 32Gb memory. The proposed method is implemented on Pytorch 1.6.0 in Python 3.7.0 environment. For a fair comparison, we adopt a consistent training hyper-parameter setting \cite{param1}. Specifically, we use SGD with Nesterov momentum for optimization and set the momentum, weight decay rate, and the initial learning rate to $0.9$, $5\times10^{-4}$ and 0.05 respectively. Each network is trained with 240 epochs, where the learning rate is scheduled to decay by 10\% at 150\textsuperscript{th}, 180\textsuperscript{th} and 210\textsuperscript{th} epoch. In the proposed framework, we set the number of peer students $J=2$, temperature $\tau=3$ for distillation, $\alpha=0.6$, $\beta=0.3$ and $\gamma=0.1$ in Eq.\eqref{eq:loss_all}. Our codes will be made publicly available after acceptance.

\subsection{Compared Methods}
We compare our proposed method with the existing state-of-the-art self knowledge distillation including Tf-KD \cite{tf-kd}, CS-KD \cite{cs-kd}, PS-KD \cite{ps-kd}, DDGSD \cite{r-aug}; and online knowledge distillation methods including DML \cite{dml}, KDCL \cite{kdcl}, ONE \cite{one} and OKDDip \cite{okddip}. For a fair comparison, we adopt two peer students in DML, and set the number of branches to 2 in KDCL, ONE and OKDDip. We follow the original settings of the remaining extra hyper-parameters. In the 'baseline', each CNN is trained directly from the hard labels. We fix the random seed at 95, and compute the average and standard deviation of the top-1 classification accuracy (\%) over three runs. 

\subsection{Experimental Results}

As shown in \tabref{table:result}, our method consistently improves
the performance of various backbone networks (baseline), achieving a 0.73-1.43\% performance gain on CIFAR-10 and 1.55-3.29\% on CIFAR-100 respectively. The fact proves the compatibility between Cut\textsuperscript{n}Mix and online distillation. And it shows its effectiveness to enhance the generalization ability of various networks. Notably, our method can also outperform state-of-the-art distillation schemes, including self-distillation and online distillation approaches. We can also make the following observations. More complex backbones, e.g. WideResNet, obtain a higher performance boost from our methods than the lightweight ones.

\subsection{Ablation Study}
Our proposed method contains three components: Cut\textsuperscript{n}Mix, MMD loss $\mathcal{L}_{MMD}$ for feature level mutual learning, and peer teacher distillation loss $\mathcal{L}_{PT}$. To analyze the effectiveness of each component, we conduct an ablation study with ResNet-110 on CIFAR-100. As illustrated in \tabref{table:ablation}, each component is also jointly effective.

\begin{table}[!ht]
\caption{
Ablation on CIFAR-100 with ResNet-110 (averaged top-1 accuracy \% over three random runs).
}\label{table:ablation}
\centering
\begin{tabular}{c|c|c|c} 
\toprule
Cut\textsuperscript{n}Mix & $\mathcal{L}_{MMD}$ & $\mathcal{L}_{PT}$ & Results \\
\hline
\ding{51} & & & 78.38\tiny{$\pm$0.25}\\
\ding{51} & \ding{51} & & 78.74\tiny{$\pm$0.13}\\
\ding{51} & & \ding{51} & 78.65\tiny{$\pm$0.08}\\
\ding{51} & \ding{51} & \ding{51} & 79.32\tiny{$\pm$0.35} \\
\bottomrule
\end{tabular}
\end{table}

\section{Conclusion}
In this research, we first empirically illustrate the orthogonality to the performance gain between the mixed sample regularization and online distillation. Furthermore, we design an even stronger mixed sample regularization named Cut\textsuperscript{n}Mix, together with a powerful online distillation framework to combine with Cut\textsuperscript{n}Mix. The proposed framework can consistently boost the classification accuracy of an ensemble of peer student networks. Due to the restriction on computational resources, evaluations on large-scale datasets such as ImageNet are left to future work. Other future directions include applications to more scenarios e.g., object detection, and image segmentation.

\bibliographystyle{IEEEbib}
\bibliography{refs}

\end{document}